# An Iterative Locally Linear Embedding Algorithm

Deguang Kong  DOOGKONG@GMAIL.COM
Chris Ding  CHQDING@UTA.EDU
Heng Huang  HENG@UTA.EDU
Feiping Nie  FEIPINGNIE@GMAIL.COM
Dept. of CSE, The University of Texas at Arlington, 500 UTA Blvd, Arlington, TX, 76019

## Abstract

Locally Linear embedding (LLE) is a popular dimension reduction method. In this paper, we systematically improve the two main steps of LLE: (A) learning the graph weights $\mathbf{W}$, and (B) learning the embedding $\mathbf{Y}$. We propose a sparse nonnegative $\mathbf{W}$ learning algorithm. We propose a weighted formulation for learning $\mathbf{Y}$ and show the results are identical to normalized cuts spectral clustering. We further propose to iterate the two steps in LLE repeatedly to improve the results. Extensive experiment results show that iterative LLE algorithm significantly improves both classification and clustering results.

## 1. Introduction

Recently, there have been many algorithms proposed for nonlinear dimension reduction, which include Isomap (Tenenbaum et al., 2000), locally linear embedding (LLE) (Roweis & Saul, 2000), kernel-LLE (Ham et al., 2004), Hessian LLE (Donoho & Grimes, 2003), local tangent alignment (Zhang & Zha, 2004), Laplacian embedding (Hall, 1971; Belkin & Niyogi, 2001), and many variations. Above dimension reduction algorithms usually cover two main steps: (A) for each data point, learn the local geometry information $\mathbf{W}$. This $\mathbf{W}$ can be viewed as similarity between data points or the edge weights of a graph whose nodes are the data points. We call this $\mathbf{W}$-learning, or learning the graph weights. (B) Using the learned $\mathbf{W}$ to embed the high-dimensional data points into a lower-dimensional space $\mathbf{Y}$. We call this $\mathbf{Y}$-learning, or learn the embedding. The performance of those algorithms are determined both by learning the local information and also by constructing the mapping relations.

In past decades, many clustering algorithms have been proposed such as K-means, spectral clustering and its variants (Ng et al., 2001), normalized cut (Shi & Malik, 1997), ratio cut (Chan et al., 1994), etc. Among them, the use of manifold information in graph cuts has shown the state-of-the-art clustering performance.

One key observation is that both LLE and spectral clustering utilize the data manifold information. This motivates us to investigate deeper relations between the LLE $\mathbf{Y}$-learning and spectral clustering in terms of Laplacian embedding (because the embedding is precisely the relaxed cluster indicators for the spectral clustering). Indeed, we discover that a properly modified formulation of $\mathbf{Y}$-learning provides a solution which is identical to the normalized Laplacian embedding (see §2.3). We incorporate this improvement into our final iterative LLE algorithm.

Another observation is that the data geometry information encoded in $\mathbf{W}$ also plays a central role in the performance of these algorithms. We investigate the $\mathbf{W}$-learning process and propose a nonnegative, kernelized, sparse $\mathbf{W}$-learning algorithm (see §4).

Furthermore, we propose to *iteratively repeat* the two mains steps ($\mathbf{W}$-learning and $\mathbf{Y}$-learning) to improve the results progressively. Here we use the learned embedding $\mathbf{Y}$ to augment the input data to learn a better $\mathbf{W}$, which leads to a better $\mathbf{Y}$ in turn. This is repeated until the process converges (details are given in §3). This iterative procedure incorporates both the improved $\mathbf{Y}$ learning and the improved $\mathbf{W}$ learning into a coherent iterative LLE algorithm.

The experiment results for clustering and semi-supervised learning tasks on 9 datasets show clear performance improvements.

## 2. LLE and New Formations

### 2.1. Brief overview of LLE

LLE (Roweis & Saul, 2000) is a nonlinear dimension reduction approach. Suppose data $\mathbf{X} = [\mathbf{x}_1, \mathbf{x}_2, \cdots, \mathbf{x}_n] \in \Re^{p \times n}$, consists of $n$ data points $\mathbf{x}_i$, each with dimensionality $p$. LLE expects each data point and its neighbors to lie on or close to a locally linear manifold, which governs how the weight coefficients $\mathbf{W}$ are constructed from Eq.(1). It then reconstruct each data point (low $k$-dimensional embedding vectors $\{\mathbf{y}_i\}$) from its neighbors via the same neighborhood relations by minimizing a quadratic cost function Eq.(2),

$$\min_{\mathbf{W}} \sum_i \|\mathbf{x}_i - \sum_{j \in \mathcal{N}_i} W_{ij} \mathbf{x}_j\|^2, \quad (1)$$

$$\min_{\mathbf{Y}} \sum_i \|\mathbf{y}_i - \sum_j W_{ij} \mathbf{y}_j\|^2, \quad (2)$$

where weight $W_{ij}$ summarizes the contribution of the $j$th data point to the construction of $i$th data point. $\mathcal{N}_i$ is the kNN neighborhood of $x_i$. The shift invariance of $\mathbf{Y} = [\mathbf{y}_1, \mathbf{y}_2, \cdots, \mathbf{y}_n] \in \Re^{k \times n}$ is enforced by restricting $\sum_j W_{ij} = 1$.

### 2.2. LLE Improvements in two directions

In this paper, we propose improved formulations in both main steps in LLE. (A) In the $\mathbf{W}$-learning step of Eq.(1), we propose new improved formulations to learn $\mathbf{W}$. We first make $\mathbf{W}$ nonnegative in this section. We will further propose a kernelized sparse learning in §4. (B) In the $\mathbf{Y}$-learning step of Eq.(2), we propose slightly modified formulation and prove that the solution to Eq.(2) is identical to Normalized Cut or Laplacian embedding. Our iterative LLE algorithm is based on these improved formations in both LLE steps.

To make a connection to graph embedding, we (1) restrict $\mathbf{W}$ to be nonnegative, i.e., we add constraint $\mathbf{W} \geq 0$ to Eq.(1) (as done in (Wang & Zhang, 2006)); (2) we symmetrize $\mathbf{W}$ to obtain $\mathbf{Z} = \frac{1}{2}(\mathbf{W} + \mathbf{W}^T)$ as the graph edge weight/similarity matrix; (3) we impose $\mathbf{D}$-orthogonal constraint on $\mathbf{Y}$, i.e., $\mathbf{Y}\mathbf{D}\mathbf{Y}^T = \mathbf{I}$, where $\mathbf{D} = \text{diag}(\mathbf{Z}e)$ is a diagonal matrix containing node degrees.

With these three changes, the LLE equations of Eqs.(1,2) become

$$\min_{\mathbf{W}} \sum_i \|\mathbf{x}_i - \sum_{j \in \mathcal{N}_i} W_{ij} \mathbf{x}_j\|^2, \quad s.t. \quad \mathbf{W} \geq 0, \quad (3)$$

$$\min_{\mathbf{Y}} \sum_i d_i \|\mathbf{y}_i - \sum_j (\mathbf{D}^{-1}\mathbf{Z})_{ij} \mathbf{y}_j\|^2 \ s.t. \ \mathbf{Y}\mathbf{D}\mathbf{Y}^T = \mathbf{I}, \quad (4)$$

where $d_i = D_{ii}$.

We note several important changes here. In Eq.(4), $\mathbf{D}^{-1}$ is inserted for two important reasons: (1) Note that $\sum_j (\mathbf{D}^{-1}\mathbf{Z})_{ij} \mathbf{y}_j$ is the average values of $\mathbf{y}_i$'s neighbors. Thus Eq.(4) enforces the smoothness of function $\{\mathbf{y}_i\}$. (2) It also enforces the shift invariance of obtained $\mathbf{Y}$, because $\sum_j (\mathbf{D}^{-1}\mathbf{Z})_{ij} = 1$. This implies that if $\{\mathbf{y}_i^*\}$ is an optimal solution, so is $\{\mathbf{y}_i^* - \mathbf{c}\}$ where $\mathbf{c}$ is a constant vector. Note that we add $d_i$ as the weight of each point $\mathbf{y}_i$, for reasons immediately clear below.

### 2.3. LLE Y-learning is identical to Normalized Cut Spectral Clustering

Now we show that LLE $\mathbf{Y}$-learning formulation of Eq.(4) is identical to normalized cut.

In fact, this is a general result, not restricted to LLE. It holds for any symmetric nonnegative graph similarity function $\mathbf{Z}$. More precisely we have theorem (1),

**Theorem 1.** *For any symmetric nonnegative graph similarity function $\mathbf{Z}$ of the formulation of Eq.(4), the optimal solution of $\mathbf{Y}$ is identical to the optimal solution $\mathbf{H}$ of normalized cut spectral clustering, given graph weight matrix $\mathbf{Z}$.*

In the following, we first briefly introduce normalized cut and present the proof of Theorem 1.

**Normalized Cut**.

Normalized cut (Shi & Malik, 1997) is an effective graph partitioning (clustering) technique to identify clusters inherent in the data, given the pairwise similarity matrix $\mathbf{Z}$. It is well-known now multi-way normalized cut can be solved by the following problem,

$$\min_{\mathbf{G}} \text{Tr}(\mathbf{G}^T(\mathbf{I} - \tilde{\mathbf{Z}})\mathbf{G}) \quad s.t. \quad \mathbf{G}^T\mathbf{G} = \mathbf{I}_k, \quad (5)$$

where $\tilde{\mathbf{Z}} = \mathbf{D}^{-\frac{1}{2}}\mathbf{Z}\mathbf{D}^{-\frac{1}{2}}$. and $\mathbf{G} = [\mathbf{g}_1, \mathbf{g}_2, \cdots, \mathbf{g}_k]$ are relaxed cluster indicators. The optimal solution for $G$ is the smallest $k$ eigenvectors from $(\mathbf{I} - \tilde{\mathbf{Z}})$, i.e.,

$$(\mathbf{I} - \tilde{\mathbf{Z}})\mathbf{g}_k = \mu_k \mathbf{g}_k. \quad (6)$$

The cluster indicator $\mathbf{H} = [\mathbf{h}_1, \mathbf{h}_2, \cdots, \mathbf{h}_k]$ is

$$\mathbf{h}_k = \mathbf{D}^{-\frac{1}{2}}\mathbf{g}_k, \qquad \mathbf{H} = \mathbf{D}^{-\frac{1}{2}}\mathbf{G}. \quad (7)$$

**Relation to Laplacian Embedding**.

It is easy to see that $\mathbf{H}^T = \mathbf{V} \equiv [\mathbf{v}_1, \cdots, \mathbf{v}_n]$ is identical to the solution of

$$\min_{\mathbf{V}} \sum_{ij} W_{ij} \|\mathbf{v}_i - \mathbf{v}_j\|^2 \quad s.t. \quad \mathbf{V}\mathbf{D}\mathbf{V}^T = \mathbf{I}_k. \quad (8)$$

This Laplacian embedding with degree normalization $\mathbf{V}\mathbf{D}\mathbf{V}^T = \mathbf{I}_k$ is effective for clustering problems because the embedding coordinates are the continuous relaxation of the cluster indicators of the multi-way normalized cut spectral clustering. Similarly, Laplacian embedding using coordinates with standard normalization $\mathbf{VV}^T = \mathbf{I}_k$ is precisely the continuous relaxation of the cluster indicators of multi-way ratio cut spectral clustering (Chan et al., 1994); The widely used linear embedding, Principal component analysis (PCA) is precisely the continuous relaxation of the cluster indicators of the multi-way K-means clustering (Zha et al., 2001; Ding & He, 2004).

Theorem 1 can be equivalently expressed for Laplacian embedding.

### 2.4. Proof of Theorem 1

To prove the theorem 1, we need Lemma 1.

**Lemma 1.** *The optimal solution to Eq.(4) is,*

$$\mathbf{Y}^* = \mathbf{F}^T \mathbf{D}^{-\frac{1}{2}}, \quad (9)$$

*where $\mathbf{F} = [\mathbf{f}_1, \mathbf{f}_2, ..., \mathbf{f}_k] \in \Re^{n \times k}$ is the smallest $k$ eigenvectors of $(\mathbf{I} - \tilde{\mathbf{Z}})^2$, $\tilde{\mathbf{Z}} = \mathbf{D}^{-\frac{1}{2}} \mathbf{Z} \mathbf{D}^{-\frac{1}{2}}$, i.e.,*

$$(\mathbf{I} - \tilde{\mathbf{Z}})^2 \mathbf{f}_k = \lambda_k \mathbf{f}_k. \quad (10)$$

**Proof of Lemma 1.**

*Proof.* Note $\mathbf{Y} = [\mathbf{y}_1, \mathbf{y}_2, \cdots, \mathbf{y}_n] \in \Re^{k \times n}$. Let

$$\tilde{\mathbf{y}}_i = \mathbf{y}_i - \sum_j (\mathbf{D}^{-1} \mathbf{Z})_{ij} \mathbf{y}_j, \quad (11)$$

and then $\tilde{\mathbf{Y}} = [\tilde{\mathbf{y}}_1, \tilde{\mathbf{y}}_2, \cdots, \tilde{\mathbf{y}}_n] \in \Re^{k \times n}$. It is easy to see $\tilde{\mathbf{Y}} = \mathbf{Y} - \mathbf{Y} \mathbf{Z} \mathbf{D}^{-1}$. Now Eq.(4) can be written as

$$\sum_{i=1}^n d_i ||\tilde{\mathbf{y}}_i||^2 = \sum_{i=1}^n \sum_{j=1}^k d_i \tilde{Y}_{ji}^2 = \sum_{j=1}^k \sum_{i=1}^n \tilde{Y}_{ji} D_{ii} (\tilde{Y}^T)_{ij}$$
$$= \text{Tr}(\tilde{\mathbf{Y}} \mathbf{D} \tilde{\mathbf{Y}}^T) = \text{Tr}\,(\mathbf{Y} - \mathbf{Y} \mathbf{Z} \mathbf{D}^{-1}) \mathbf{D} (\mathbf{Y} - \mathbf{Y} \mathbf{Z} \mathbf{D}^{-1})^T$$
$$= \text{Tr}\,\mathbf{Y}(\mathbf{I} - \mathbf{Z} \mathbf{D}^{-1}) \mathbf{D}^{\frac{1}{2}} \mathbf{D}^{\frac{1}{2}} [\mathbf{Y}(\mathbf{I} - \mathbf{Z} \mathbf{D}^{-1})]^T$$
$$= \text{Tr}\,\mathbf{Y} \mathbf{D}^{\frac{1}{2}} (\mathbf{I} - \tilde{\mathbf{Z}})(\mathbf{I} - \tilde{\mathbf{Z}}) \mathbf{D}^{\frac{1}{2}} \mathbf{Y}^T.$$

Thus Eq.(4) becomes

$$\min_{\mathbf{Y}} \text{Tr}(\mathbf{Y} \mathbf{D}^{\frac{1}{2}} (\mathbf{I} - \tilde{\mathbf{Z}})^2 \mathbf{D}^{\frac{1}{2}} \mathbf{Y}^T) \quad s.t. \quad \mathbf{Y} \mathbf{D} \mathbf{Y}^T = \mathbf{I}. \quad (12)$$

To optimize Eq.(12) is equivalent to optimize,

$$\min_{\mathbf{F}} \text{Tr} \mathbf{F}^T (\mathbf{I} - \tilde{\mathbf{Z}})^2 \mathbf{F}, \quad s.t. \quad \mathbf{F}^T \mathbf{F} = \mathbf{I}, \quad (13)$$

where $\mathbf{F} = \mathbf{D}^{\frac{1}{2}} \mathbf{Y}^T$. It is easy to see the optimal solution $\mathbf{F} = [\mathbf{f}_1, \mathbf{f}_2, \cdots, \mathbf{f}_k]$ for Eq.(13) is the smallest $k$ eigenvectors from $(\mathbf{I} - \tilde{\mathbf{Z}})^2$, i.e., Eq.(9). Thus the optimal solution $\mathbf{Y}^* = (\mathbf{D}^{-\frac{1}{2}} \mathbf{F})^T = \mathbf{F}^T \mathbf{D}^{-\frac{1}{2}}$. □

**Proof of Theorem 1.**

*Proof.* Because $(\mathbf{I} - \tilde{\mathbf{Z}})$ is semi-definite positive, the eigenvectors $\mathbf{g}_k$ of Eq.(6) can be uniquely mapped to eigenvectors $\mathbf{g}_k$ of

$$(\mathbf{I} - \tilde{\mathbf{Z}})^2 \mathbf{g}_k = \mu_k^2 \mathbf{g}_k. \quad (14)$$

Comparing Eq.(6) of normalized cut against Eq.(10) of LLE, one can see $\mathbf{f}_i = \mathbf{g}_i, \mu_i^2 = \lambda_i, \mathbf{F} = \mathbf{G}$. Compared Eq.(7) of normalized cut against Eq.(9) of LLE, one can see $\mathbf{H} = \mathbf{Y}^T$. This completes the proof. □

## 3. An Iterative LLE Learning Algorithm (ILLE)

We now use the above results, coupled with two new schemes(A,B) to derive a new learning algorithm.

### 3.1. Motivation of iterative LLE

(A)**Iterative process of LLE**

In LLE, starting from $\mathbf{X}$, we learn $\mathbf{W}$, and then learn $\mathbf{Y}$ as the low-dimensional embedding of data $\mathbf{X}$. In this paper, we propose to use $\mathbf{Y}$ as the new data and iterate this process to further improve the embedding. The key observation is that the class structure of the data is more clear in $\mathbf{Y}$ than in $\mathbf{X}$ (this is the original embedding purpose of LLE). Thus we use $\mathbf{Y}$ as the new data and repeat this process to learn an improved $\mathbf{Y}$.

(B) **Kernel generalization**

From experiments on several datasets, the results of using linear formulation on $\mathbf{X}$ for learning $\mathbf{W}$ in Eq.(1) are generally not as good as other state-of-art methods. Here we use the kernel trick to generalize this to arbitrary nonlinear similarity function. We re-write Eq.(1) as

$$\min_{\mathbf{W}} \sum_i ||\phi(\mathbf{x}_i) - \sum_{j \in \mathcal{N}_i} \mathbf{W}_{ij} \phi(\mathbf{x}_i)||^2, \quad (15)$$

where $\phi(\mathbf{x}_i)$ is a mapping to a higher dimensional space. The important thing here is that the exact form of the mapping function is not needed; only the inner product $\mathcal{K}_{ij} = \langle \phi(\mathbf{x}_i), \phi(\mathbf{x}_j) \rangle$ is needed.

Using matrix notation, the LLE of Eq.(1) can be written as $\min_{\mathbf{W}} ||\mathbf{X} - \mathbf{X} \mathbf{W}^T||^2$, and Eq.(15) can be written as

$$||\phi(\mathbf{X}) - \phi(\mathbf{X}) \mathbf{W}^T||^2 = \text{Tr}(\mathcal{K} - \mathbf{W} \mathcal{K} - \mathcal{K} \mathbf{W}^T + \mathbf{W} \mathcal{K} \mathbf{W}^T). \quad (16)$$

This is useful, because once we compute $\mathbf{Y}$ from Eq.(4), we can build a kernel from $\mathbf{Y}$ and substitute it into Eq.(16) to learn a new $\mathbf{W}$ (and thus $\mathbf{Z}$).

## 3.2. Proposed algorithm

By incorporating the above schemes of (A,B), we outline our iterative LLE learning algorithm as follows.

(1) Given kernel $\mathcal{K}^t$, solve for $\mathbf{W}^t$ with Eq.(16) or Eq.(18)[1].

(2) Given pairwise similarity $\mathbf{W}^t$, solve for $\mathbf{Y}^t$ using Lemma 1.

(3) Given embedding $\mathbf{Y}^t$, compute a new kernel $\mathcal{K}^{t+1}$ either as the final result of our algorithm (both embedding $\mathbf{Y}^t$ and kernel $\mathcal{K}^{t+1}$) or as input to step (1). Details of $\mathcal{K}^{t+1}$ construction is given in §3.3.

Initially $\mathcal{K}^1$ is obtained from data $\mathbf{X}$, we repeat above 3 steps for serval iterations to obtain a better kernel. See Algorithm 1 for more details. Note in step(1), we have two alternatives to compute $\mathbf{W}^t$. Thus we have two versions of iterative LLE - one based on simply iterating LLE process, and the other based on learning a sparse kernel using algorithm of Eq.(21) in §4.

**Discussion** Here we did not give the global convergence proof of this iterative LLE algorithm. The algorithm is very intuitive and natural. It is motivated by a simple observation: class structure is more clear in embedding $\mathbf{Y}$ than in original data $\mathbf{X}$.

### 3.3. Construction of the new kernel

In step (3) of our algorithm, once the low-dimensional embedding $\mathbf{Y}^t$ is obtained, we have the following choices.

(a) Construct a new kernel from $\mathbf{Y}^t$. There are many way to construct kernel. One possible approach is to construct the kernel $\mathcal{K}_\mathbf{Y}$ by simply using the Gaussian Kernel, i.e., $\mathcal{K}_\mathbf{Y} = e^{-\gamma ||\mathbf{y}_i - \mathbf{y}_j||^2}$, where $\gamma$ is the scale parameter. Another way is to construct new kernel $\mathcal{K}_\mathbf{Y}$ as the linear kernel in low-dimensional space, i.e., $\mathcal{K}_\mathbf{Y} = \mathbf{Y}\mathbf{Y}^T$.

(b) Construct the kernel $\mathcal{K}^{t+1}$ either as the final result of our algorithm or as input to step (1). There are many choices, (b1) $\mathcal{K}^{t+1} = \mathcal{K}_\mathbf{Y}^t$; (b2) Kernel $\mathcal{K}^{t+1}$ is a combination of $\mathcal{K}_\mathbf{Y}^t$ and the previous kernel $\mathcal{K}^t$. There are two way to achieve this, additively $\mathcal{K}^{t+1} = \mathcal{K}^t + \mathcal{K}_\mathbf{Y}^t$, or multiplicatively $\mathcal{K}^{t+1} = \mathcal{K}^t \odot \mathcal{K}_\mathbf{Y}^t$, where we use $\odot$ to denote the element-wise matrix multiplication, e.g., if $\mathbf{C} = \mathbf{A} \odot \mathbf{B}$, then $C_{ij} = A_{ij} \times B_{ij}$.

In choice (b1), we simply ignore the previous kernel and set the new kernel $\mathcal{K}^{t+1} = \mathcal{K}_\mathbf{Y}^t$. Note both additive and multiplicative operations in choices (b2) ensure the new kernel $\mathcal{K}^{t+1}$ is also semi-definite positive(s.d.p) if the original kernel $\mathcal{K}^t$ is s.d.p.

**Discussion** In our experiments, we tried different choices. We find the results obtained from (b2) are generally better than (b1), and the multiplicative combination usually achieves better results than additive combination. Thus in our experiment we use (b2) with multiplicative combination to construct the new kernel in step 3.

---

**Algorithm 1** Iterative LLE algorithm(ILLE)

**Input:** Original Kernel $\mathcal{K}^1$ obtained from data $\mathbf{X}$, maximal iteration $T$
**Output:** Pairwise similarity $\mathbf{W}$, embedding $\mathbf{Y}$
**Algorithm:**
1: **for** $t = 1$ to $T$ **do**
2:   Compute $\mathbf{W}^t$ of Eq.(16) or Eq.(18) with current kernel $\mathcal{K}^t$
3:   Compute $\mathbf{Z}^t = \frac{1}{2}(\mathbf{W} + \mathbf{W}^t)$.
4:   Compute embedding $\mathbf{Y}^t$ using Lemma 1.
5:   Compute a new kernel $\mathcal{K}^{t+1}$ given embedding $\mathbf{Y}^t$.
6: **end for**
7: **Output**: Pairwise similarity $\mathbf{W} = \mathcal{K}^{t+1}$, embedding $\mathbf{Y} = \mathbf{Y}^t$.

---

## 4. Improved W-Learning Formulation

Here we propose an improvement to the $\mathbf{W}$-learning step of LLE. So far for LLE of Eq.(1) and the new kernel version of Eq.(16), we maintain the original LLE convention that $\mathbf{W}$ preserves the $k$NN structure, i.e. $W_{ij} \neq 0$ for only $j \in \mathcal{N}_i$ ($k$NN of object $i$).

This constraint is too strong for constructing the data similarity matrix $\mathbf{W}$. Thus, in our approach, we relax this to let $\mathbf{W}_{ij}$ be nonzero even if $j \notin \mathcal{N}_i$. In other words, we bypass kNN entirely.

We now present a new approach to learn the pairwise similarity matrix $\mathbf{S} \in \Re^{n \times n}$, where $\mathbf{S}_{ij}$ represents the $i$-th data's contribution to reconstruct data point $x_j$. We hope the newly learned $\mathbf{S}$ has much clear structure. We use the symbol $\mathbf{S}$ to emphasize that $\mathbf{W}$ is learned using the new approach. Our objective function for learning $\mathbf{S}$ is,

$$\min_{\mathbf{S} \geq 0} ||\mathbf{X} - \mathbf{XS}||^2 + \alpha \text{Tr}(\mathbf{S}^T\mathbf{S}) + \beta ||\mathbf{S}||_{1,1}, \quad (17)$$

where $\alpha$ and $\beta$ are regularization parameters, $||\mathbf{S}||_{1,1} = \sum_{ij} |\mathbf{S}_{ij}|$. The first term $||\mathbf{X} - \mathbf{XS}||^2 = \sum_i ||\mathbf{x}_i - \sum_j \mathbf{S}_{ji}\mathbf{x}_j||^2$ is used to minimize the reconstruction error from the original data. The second term penalizes the complexity of $\mathbf{S}$. The third term of $L_1$ norm is to promote the sparsity of the solution.

Using mapping $\phi: \mathbf{X} \to \phi(\mathbf{X})$ to map data $\mathbf{X}$ to a higher dimensional space in kernel machine. Eq.(17)

---

[1] $\mathbf{S}$ of Eq.(18) can be viewed as pairwise similarity

An Iterative Locally Linear Embedding Algorithm

becomes

$$\min_{\mathbf{S}\geq 0} \ \|\phi(\mathbf{X}) - \phi(\mathbf{X})\mathbf{S}\|^2 + \alpha \mathrm{Tr}(\mathbf{S}^T\mathbf{S}) + \beta \|\mathbf{S}\|_{1,1}, \quad (18)$$

which is equivalent to,

$$\min_{\mathbf{S}\geq 0} \ \mathrm{Tr}(\mathcal{K} - 2\mathcal{K}\mathbf{S} + \mathbf{S}^T\mathcal{K}\mathbf{S}) + \alpha \mathrm{Tr}(\mathbf{S}^T\mathbf{S}) + \beta \|\mathbf{S}\|_{1,1}. \quad (19)$$

Eq.(19) is identical to Eq.(17) when $\mathcal{K} = \mathbf{X}^T\mathbf{X}$.

Eq.(19) is a convex optimization problem and $\mathbf{S}$ has a unique global solution. Furthermore, Eq.(19) can be written as

$$\min_{\mathbf{S}\geq 0} \ \mathrm{Tr}[\mathcal{K} + (\beta \mathbf{E} - 2\mathcal{K})\mathbf{S} + \mathbf{S}^T(\mathcal{K} + \alpha \mathbf{I})\mathbf{S}], \quad (20)$$

where $\mathbf{E}$ is a matrix of all ones. Because $\mathcal{K}$ is s.d.p., by adding $\alpha \mathbf{I}$ with $\alpha > 0$, $(\mathcal{K} + \alpha \mathbf{I})$ is a well-conditioned matrix. It can be solved efficiently (see below). Usually $L_1$ norm term is difficult to handle. Here, however, it does not add any difficulty when handled together with the nonnegativity constraint. The $L_1$ term can be ignored entirely: $\|\mathbf{S}\|_{1,1} = \mathrm{Tr}(\mathbf{ES})$.

### 4.1. Computational algorithm for Eq.(18)

Here we present an efficient algorithm to solve Eq.(18) and prove its convergence rigorously.

The algorithm starts with an initial guess of $\mathbf{S} = \mathbf{E}$ ($\mathbf{E}$ is a matrix of all ones), iteratively updates $\mathbf{S}$ according to

$$\mathbf{S}_{ij} \leftarrow \mathbf{S}_{ij} \frac{\mathcal{K}_{ij}}{(\mathcal{K}\mathbf{S} + \alpha\mathbf{S})_{ij} + \frac{\beta}{2}}. \quad (21)$$

This algorithm converges very fast. The computational algorithm for Eq.(18) is very simple and can be efficiently implemented.

### 4.2. Convergence of Updating rule of Eq.(21)

We have Theorem(2) to prove the convergence of the algorithm when $\mathcal{K}$ is non-negative.

**Theorem 2.** *Updating $\mathbf{S}$ using the rule of Eq.(21), the objective function of Eq.(18) monotonically decreases.*

The proof of this theorem is lengthy and is similar to that in (Ding et al., 2010; Kong et al., 2011). We therefore skip the proof in this paper.

### 4.3. Correctness of Updating Rule of Eq.(21)

We prove that the converged solution satisfies the Karush-Kuhn-Tucker condition of the constrained optimization theory. We have Theorem 3 to prove it.

**Theorem 3.** *At convergence, the converged solution $\mathbf{S}$ of the updating rule of Eq.(21) satisfies the KKT condition of the optimization theory.*

Table 1. Dataset descriptions.

| Dataset | #Size | #Dimension | #Class |
|---|---|---|---|
| AT&T | 400 | 644 | 40 |
| Mnist | 150 | 784 | 10 |
| Umist | 360 | 644 | 20 |
| Binalpha | 1014 | 320 | 36 |
| Yale | 1984 | 2016 | 31 |
| Caltec | 600 | 432 | 20 |
| MSRC | 210 | 432 | 7 |
| Newsgroup | 499 | 500 | 5 |
| Reuters | 900 | 1000 | 10 |

*Proof.* The KKT condition for $\mathbf{S}$ with constraints $\mathbf{S}_{ij} \geq 0$ is $\frac{\partial J(\mathbf{S})}{\partial \mathbf{S}_{ij}}\mathbf{S}_{ij} = 0$, $\forall \ i,j$.

The derivative of $J(\mathbf{S})$(Eq.18) is $\frac{\partial J(\mathbf{S})}{\partial \mathbf{S}_{ij}} = (-2\mathcal{K} + 2\mathcal{K}\mathbf{S} + 2\alpha\mathbf{S} + \beta\mathbf{E})_{ij}$. Thus the KKT condition for $\mathbf{S}$ is

$$(-2\mathcal{K} + 2\mathcal{K}\mathbf{S} + 2\alpha\mathbf{S} + \beta\mathbf{E})_{ij}\mathbf{S}_{ij} = 0 \quad \forall \ i,j. \quad (22)$$

On the other hand, once $\mathbf{S}$ converges, according to the updating rule of Eq.(21), the converged solution $\mathbf{S}$ satisfies

$$\mathbf{S}_{ij} = \mathbf{S}_{ij}\frac{\mathcal{K}_{ij}}{(\mathcal{K}\mathbf{S} + \alpha\mathbf{S} + \frac{\beta}{2}\mathbf{E})_{ij}}, \quad (23)$$

which can be written as $[-\mathcal{K}_{ij} + (\mathcal{K}\mathbf{S} + \alpha\mathbf{S} + \frac{\beta}{2}\mathbf{E})_{ij}]\mathbf{S}_{ij} = 0$. This is identical to Eq.(22). Thus the converged solution satisfies the KKT condition. □

## 5. Experiments

We perform the proposed algorithms on nie datasets. We do both semi-supervised learning and clustering on these datasets. We evaluate the proposed iterative LLE learning algorithm(§3) and sparse similarity learning algorithm(§4), and then show the embedding results from our approach.

**Dataset** These data sets come from a wide range of domains, including three face datasets AT&T, umist and yale (Georghiades et al., 2001), two digit datasets mnist (Lecun et al., 1998) and binalpha [1], two image scene datasets Caltec101(Caltec) (Dueck & Frey, 2007) and MSRC (Lee & Grauman, 2009), and two text datasets Newsgroup[2], Reuters[3]. Table 1 summarizes the characteristics of them.

---

[1] http://www.kyb.tuebingen.mpg.de/ssl-book/ benchmarks.html
[2] http://people.csail.mit.edu/jrennie/20Newsgroups/
[3] http://www.daviddlewis.com/resources/testcollections/reuters21578/

Table 2. Accuracy(ACC), normalized mutual information (NMI), and purity(PUR) comparisons of different clustering algorithms: Normalized Cut, Symmetric NMF and Spectral Clustering. $\mathcal{K}^0$: results obtained on the original/input kernel. LLE1: results on learned $\mathbf{Y}$ after 1 LLE iteration. LLE4: results on learned $\mathbf{Y}$ after 4 LLE iterations. All results shown are percentage.

| Dataset | Metric | Normalized Cut | | | Symmetric NMF | | | Spectral Clustering | | |
|---|---|---|---|---|---|---|---|---|---|---|
| | | $\mathcal{K}^0$ | LLE1 | LLE4 | $\mathcal{K}^0$ | LLE1 | LLE4 | $\mathcal{K}^0$ | LLE1 | LLE4 |
| AT&T | ACC | 44.77 | 50.24 | 66.50 | 48.09 | 49.12 | 50.04 | 41.09 | 53.18 | 58.31 |
| | NMI | 70.14 | 74.23 | 83.82 | 62.28 | 65.87 | 70.51 | 59.40 | 68.50 | 74.67 |
| | PUR | 49.30 | 54.87 | 71.49 | 48.33 | 50.32 | 54.78 | 48.00 | 49.24 | 52.41 |
| Mnist | ACC | 64.37 | 64.87 | 65.61 | 73.29 | 76.43 | 81.84 | 73.29 | 74.21 | 75.14 |
| | NMI | 65.77 | 66.84 | 67.25 | 69.83 | 72.03 | 74.92 | 73.03 | 73.38 | 74.93 |
| | PUR | 66.55 | 67.12 | 68.37 | 74.16 | 76.87 | 81.88 | 74.69 | 74.89 | 75.61 |
| Umist | ACC | 48.44 | 48.85 | 49.11 | 49.46 | 49.87 | 50.24 | 43.13 | 44.87 | 45.76 |
| | NMI | 64.62 | 64.98 | 65.15 | 64.56 | 65.34 | 66.95 | 63.26 | 63.78 | 63.89 |
| | PUR | 52.06 | 52.92 | 53.71 | 52.43 | 53.14 | 54.98 | 48.85 | 49.23 | 50.72 |
| Binalpha | ACC | 40.52 | 42.23 | 45.91 | 40.65 | 42.78 | 44.67 | 39.18 | 42.45 | 44.26 |
| | NMI | 56.25 | 57.65 | 60.35 | 54.49 | 55.61 | 59.54 | 53.57 | 56.72 | 58.51 |
| | PUR | 43.58 | 45.54 | 49.57 | 43.60 | 45.71 | 48.73 | 41.82 | 45.23 | 48.07 |
| Yale | ACC | 9.02 | 12.21 | 15.49 | 10.72 | 11.34 | 14.78 | 10.83 | 10.98 | 12.89 |
| | NMI | 11.24 | 13.43 | 20.12 | 13.98 | 16.84 | 20.45 | 12.72 | 13.45 | 16.58 |
| | PUR | 9.93 | 15.53 | 16.57 | 11.71 | 13.23 | 15.69 | 11.72 | 12.37 | 13.76 |
| Caltec | ACC | 36.31 | 42.43 | 49.51 | 43.98 | 47.83 | 52.50 | 43.67 | 45.74 | 47.98 |
| | NMI | 42.63 | 45.45 | 54.86 | 48.25 | 52.01 | 56.43 | 48.02 | 50.23 | 51.84 |
| | PUR | 39.02 | 42.58 | 53.18 | 46.21 | 50.38 | 55.71 | 46.41 | 49.65 | 51.32 |
| MSRC | ACC | 53.23 | 60.89 | 66.65 | 57.86 | 62.34 | 66.77 | 65.85 | 66.78 | 68.42 |
| | NMI | 44.08 | 50.23 | 55.81 | 46.81 | 49.87 | 56.16 | 54.78 | 55.23 | 56.36 |
| | PUR | 55.89 | 61.43 | 69.95 | 60.12 | 64.23 | 69.62 | 67.38 | 68.84 | 69.64 |
| Newsgroup | ACC | 27.58 | 32.23 | 40.36 | 26.62 | 34.78 | 51.63 | 42.22 | 44.38 | 46.51 |
| | NMI | 12.92 | 18.24 | 19.41 | 17.65 | 27.86 | 30.22 | 18.01 | 20.32 | 23.19 |
| | PUR | 28.43 | 32.54 | 41.95 | 29.12 | 42.45 | 59.20 | 41.72 | 44.81 | 48.90 |
| Reuters | ACC | 19.22 | 23.87 | 30.59 | 24.02 | 35.98 | 41.35 | 33.48 | 34.49 | 35.83 |
| | NMI | 15.69 | 18.42 | 22.22 | 11.30 | 26.83 | 32.74 | 24.26 | 25.80 | 27.78 |
| | PUR | 19.97 | 23.34 | 33.39 | 24.98 | 31.90 | 45.92 | 37.91 | 37.98 | 38.43 |
| **Average** | ACC | 38.16 | 41.98 | 47.75 | 41.63 | 45.61 | 50.42 | 43.64 | 46.34 | 48.34 |
| | NMI | 42.59 | 45.50 | 49.89 | 43.24 | 48.03 | 52.00 | 45.23 | 47.49 | 49.75 |
| | PUR | 40.53 | 43.99 | 50.91 | 43.41 | 47.58 | 54.06 | 46.50 | 48.03 | 49.87 |

We show both the iterative LLE (algorithm 1 in §3) and the sparse similarity learning algorithm (§4) results. Given original kernel $\mathcal{K}^0$, $\mathbf{S}$ is obtained from 1-time running of sparse similarity learning algorithm in §4. Then we obtain the final embedding results $\mathbf{Y}$ after repeating out iterative LLE algorithm for 4 times. For step 1 of algorithm 1(§3), we use kernel constructed from Eq.(18) for the subsequent iterations. For step 3 of algorithm 1(§3), given current embedding $\mathbf{Y}^t$, we obtain the new kernel $\mathcal{K}^{t+1}$ using choice (b2) with multiplicative combination in every iteration.

### 5.1. Clustering Results

We use clustering algorithms to evaluate the learned $\mathbf{Y}$ in LLE. We compare three standard clustering algorithms: (1) normalized cut, which in the context of our iterative LLE, is simply K-means clustering on learned embedding $\mathbf{Y}$; (2) spectral clustering (Ng et al., 2001), which is K-means clustering on embedding $\mathbf{Y}$ normalized onto unit sphere. (3) symmetric NMF, which runs on the learned $\mathbf{W}$ in iterative LLE. All of results are the averages of 10 K-means clustering with random starts.

We use accuracy, normalized mutual information (NMI) and purity as the measurement of the clustering qualities and the results are shown in Table 2. We show the clustering results obtained from using (1) the original/input kernel ($\mathcal{K}^0$), (2) LLE1: results on learned $\mathbf{Y}$ after 1 LLE iteration. (3) LLE4: results on learned $\mathbf{Y}$ after 4 LLE iterations.

For image datasets, we use gaussian kernel $\mathcal{K}^0_{ij} = e^{-\gamma||\mathbf{x}_i-\mathbf{x}_j||^2}$. For text datasets, we use linear kernel. We tune the graph construction parameter $\gamma$ to obtain the best results from kernel $\mathcal{K}^0$. From Table 2, we observe that LLE1 and LLE4 consistently achieve better clustering results, as compared to the results obtained from original kernel $\mathcal{K}^0$.

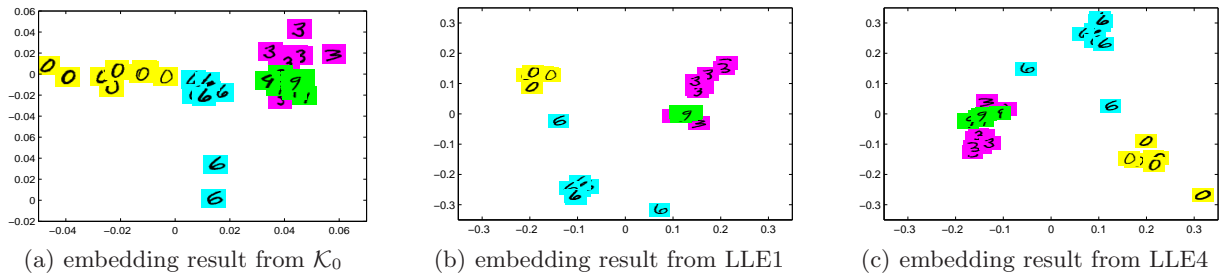

(a) embedding result from $\mathcal{K}_0$    (b) embedding result from LLE1    (c) embedding result from LLE4

Figure 1. 2D visualizations of embedding results using (1) initial/input kernel $\mathcal{K}_0$; (2) LLE1: results on learned **Y** after 1 LLE iteration; (3) LLE4: results on learned **Y** after 4 LLE iterations; using 4 digits "0","3","6","9" on MNIST dataset

Table 3. Accuracy comparisons of semi-supervised learning on 9 datasets. Learning algorithms used: Harmonic function, Green's function and Local and global consistency(LG-consistency). $\mathcal{K}^0$: results obtained on the original/input kernel. LLE1: results on learned **W** after 1 LLE iteration. LLE4: results on learned **W** after 4 LLE iterations. Results shown are based on 10% or 20% labeled data.

| Dataset | Percent-labeled | Harmonic function | | | Green's function | | | LG-consistency | | |
|---|---|---|---|---|---|---|---|---|---|---|
| | | $\mathcal{K}^0$ | LLE1 | LLE4 | $\mathcal{K}^0$ | LLE1 | LLE4 | $\mathcal{K}^0$ | LLE1 | LLE4 |
| AT&T | 10% | 65.63 | 70.23 | 73.14 | 69.67 | 70.12 | 71.11 | 70.48 | 71.45 | 72.12 |
| | 20% | 74.93 | 78.87 | 83.37 | 78.01 | 79.03 | 79.73 | 78.43 | 80.23 | 82.94 |
| Mnist | 10% | 68.83 | 68.89 | 69.91 | 63.35 | 63.90 | 64.21 | 65.72 | 67.89 | 69.19 |
| | 20% | 81.16 | 82.09 | 82.83 | 72.67 | 73.42 | 74.16 | 75.51 | 79.38 | 81.33 |
| Umist | 10% | 48.64 | 50.45 | 51.19 | 47.91 | 48.03 | 48.42 | 48.87 | 49.35 | 50.68 |
| | 20% | 63.78 | 67.89 | 70.43 | 60.75 | 61.23 | 61.54 | 63.28 | 68.78 | 70.48 |
| Binalpha | 10% | 47.71 | 49.89 | 52.61 | 46.79 | 47.09 | 49.24 | 46.76 | 48.93 | 50.35 |
| | 20% | 53.51 | 59.23 | 61.78 | 52.70 | 53.28 | 54.34 | 52.59 | 59.37 | 61.21 |
| Yale | 10% | 30.31 | 35.43 | 38.54 | 29.13 | 31.99 | 32.94 | 34.67 | 37.65 | 43.23 |
| | 20% | 45.48 | 52.45 | 54.18 | 32.09 | 33.45 | 36.55 | 38.98 | 48.90 | 57.49 |
| Caltec | 10% | 44.46 | 48.76 | 54.38 | 44.79 | 45.08 | 45.24 | 44.52 | 48.75 | 53.64 |
| | 20% | 49.87 | 53.25 | 63.67 | 49.03 | 50.23 | 52.34 | 49.93 | 53.74 | 63.62 |
| MSRC | 10% | 57.46 | 60.35 | 66.50 | 59.47 | 60.01 | 60.24 | 60.12 | 63.45 | 65.82 |
| | 20% | 62.26 | 65.43 | 70.95 | 61.42 | 62.23 | 63.54 | 63.33 | 68.79 | 72.15 |
| Newsgroup | 10% | 65.16 | 67.34 | 69.85 | 53.35 | 54.23 | 55.47 | 56.39 | 57.78 | 58.37 |
| | 20% | 72.27 | 73.25 | 74.35 | 59.72 | 60.91 | 61.14 | 58.84 | 60.19 | 61.32 |
| Reuters | 10% | 64.25 | 65.78 | 66.23 | 53.29 | 55.79 | 57.81 | 53.27 | 58.98 | 61.44 |
| | 20% | 73.61 | 73.98 | 74.56 | 62.35 | 63.45 | 68.74 | 61.09 | 67.90 | 72.17 |
| **Average** | 10% | 54.72 | 57.46 | 60.26 | 51.97 | 52.92 | 53.85 | 53.42 | 56.03 | 58.32 |
| | 20% | 64.10 | 67.38 | 70.68 | 58.75 | 59.69 | 61.34 | 60.22 | 65.25 | 69.19 |

### 5.2. Semi-supervised learning results

We use $\mathcal{K}^0$, LLE1 and LLE4 results (learned **W**) as the input to run three semi-supervised methods: harmonic function(Zhu et al., 2003), local and global consistency(Zhou et al., 2004), green's function(Ding et al., 2007). We compare the classification accuracy of above three methods by using original kernel($\mathcal{K}^0$) and the results obtained from LLE1 and LLE4 on 9 data sets. For all the methods and datasets, we randomly select 10%, 20% of labeled data for each class, and use the rest as unlabeled data. We do 10 fold and 5 fold cross validation, respectively. Finally, we report the average of the semi-supervised classification accuracy in Table 3. In all cases, we obtain higher classification accuracy by applying iterative LLE learning algorithm (shown as LLE4 and LLE1).

### 5.3. Demonstration of embedding results

We demonstrate the advantages of iterative LLE learning algorithm (§3) and sparse similarity learning algorithm (§4) using two-dimensional visualization. We randomly select four digits from MNIST dataset ("0", "3", "6", "9"). Given Gaussian Kernel as the input, the iterative LLE algorithm (§3) and sparse similarity learning algorithm (§4) are run. The other parameters are set as mentioned before. The embedding results obtained from original Gaussian Kernel $\mathcal{K}_0$, 4-time running of iterative LLE learning algorithm (LLE4) and 1-time running of **W**-learning algorithm (LLE1) are shown in Figs.(1(a),1(c),1(b)). In original results from Gaussian Kernel, all images from different groups collapse together. For the results obtained from LLE4 and LLE1, the images from different groups are balanced and distributed more evenly. This indicates much better embedding results.

**Insights from experiment results**. Overall, from initial/input kernel $\mathcal{K}^0$ to LLE1, LLE4, both clustering and semi-supervised learning results consistently improved. Comparing results obtained between LLE1 and initial/input kernel $\mathcal{K}^0$, the performance boost is from the learned **W** using the algorithm of §4. Comparing results obtained between LLE1 and LLE4, the performance boost is from the iterative learning of LLE. From the statistics shown in Tables 2, 3, we observe that the boost from LLE1 to LLE4 is usually higher than that from $\mathcal{K}^0$ to LLE1, indicating that the iterative aspect contributes more.

## 6. Conclusion

In summary, the main contribution of our paper is in three-fold. (1) We show that an improved **Y**-learning formulation of LLE is identical to normalized cut spectral clustering. (2) We present an improved **W**-learning algorithm that learns a nonnegative, sparse pairwise similarity from an input kernel function. (3) An iterative procedure of the above two steps is proposed to progressively refine/improve the solution. Experiments show that the iterative LLE incorporating (1,2,3) leads to better clustering and semi-supervised learning results.

**Acknowledgments**. This work is supported partially by NSF-CCF-0939187, NSF-CCF-0917274, NSF-DMS-0915228.